\documentclass[journal,twoside,web]{ieeecolor}
\usepackage{generic}
\usepackage{cite}
\usepackage{amsmath,amssymb,amsfonts}
\usepackage{algorithmic}
\usepackage{graphicx}
\usepackage{textcomp}
\def\BibTeX{{\rm B\kern-.05em{\sc i\kern-.025em b}\kern-.08em
    T\kern-.1667em\lower.7ex\hbox{E}\kern-.125emX}}
\usepackage{graphicx} 
\usepackage{epsfig}
\usepackage{epstopdf} 

\usepackage{booktabs} 
\usepackage{multirow}
\usepackage{url}
\newcommand{\cmdSOL}{\texttt{SOL}}
\newcommand{\cmdEOS}{\texttt{EOS}}
\newcommand{\cmdLine}{\texttt{Line}}
\newcommand{\cmdArc}{\texttt{Arc}}
\newcommand{\cmdCirc}{\texttt{Circle}}
\newcommand{\cmdExt}{\texttt{Extrude}}

\renewcommand{\footnoterule}{%
  \kern -3pt                    % 向上微调线条位置，防止与正文离得太近
  \hrule width 0.25\textwidth height 0.4pt % 线条设置：宽度为版面宽度的 1/4，粗细为 0.4pt
  \kern 2.6pt                   % 向下微调位置，保持与脚注文本的间距
}

\begin{document}
\title{VGGT-CAD: Reconstructing Parametric CAD 3D Model with Geometric Grounding}

\author{Chunan Yu, Tianrun Chen, Fu Shen, Cheng Chen, Lanyun Zhu, Yang Yang \\}

\twocolumn[{
\renewcommand\twocolumn[1][]{#1}
\maketitle
\begin{center}
\vspace{-1cm}
    \includegraphics[width=\textwidth]{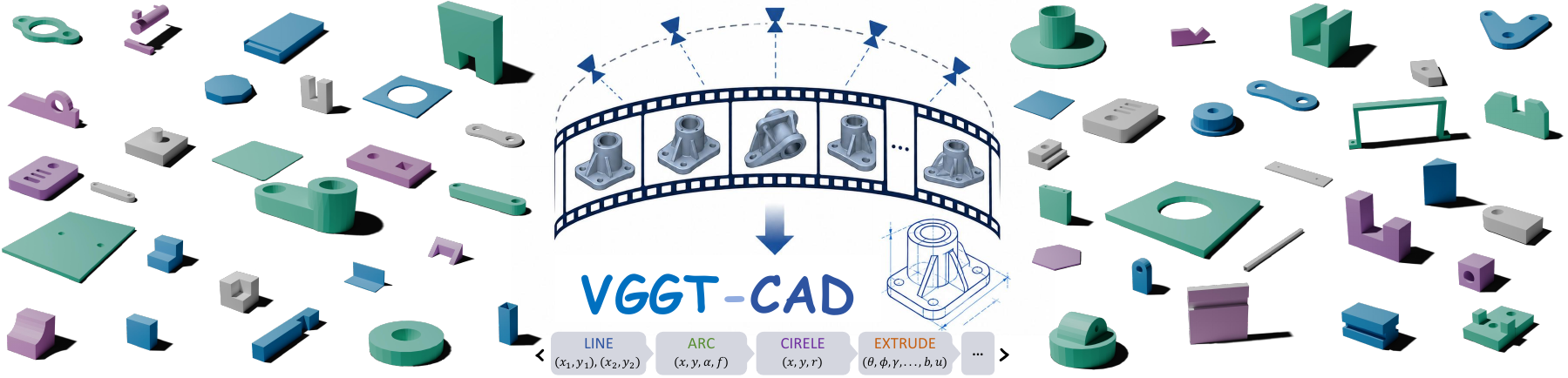}
\end{center}
\noindent
\refstepcounter{figure} % Increment the figure counter
\footnotesize \textsf{Fig. \thefigure. \hspace{0.3em} 
Given unconstrained single-view images or multi-view videos of a target object, our VGGT-CAD framework infers editable parametric CAD command sequences and 3D solid models with high geometric fidelity in a feed-forward manner.}
}
\vspace{+2mm}
]

\renewcommand{\thefootnote}{}
\footnotetext{This work was supported by the National Natural Science Foundation of China (No. 62276131), the Natural Science Foundation of Jiangsu Province (No. BK20240081) and the Special Research Project on Teaching Reform of General Artificial Intelligence Courses in Jiangsu Undergraduate Universities (No. ZNT-10). (Corresponding author: Yang Yang.)}
\footnotetext{Chunan Yu and Yang Yang are with the School of Computer Science and Engineering, Nanjing University of Science and Technology, Nanjing 210094, China (e-mail: \{yyang, yuchun\_an\}@njust.edu.cn).}
\footnotetext{Tianrun Chen, Chen Cheng and Lanyun Zhu are with the KOKONI3D, Moxin (Huzhou) Technology Co., Ltd., China (e-mail: tianrun.chen@zju.edu.cn).}
\footnotetext{Fu Shen is with the Nanjing Institute of Agricultural Mechanization, Ministry of Agriculture and Rural Affairs, Nanjing 210014, China.}

\begin{abstract}
    Parametric CAD reconstruction requires recovering both precise geometry and editable modeling operations from visual observations, making it challenging under limited and ambiguous views. Existing methods mainly rely on 2D appearance cues and lack strong multi-view geometric priors. In this work, we present \textbf{VGGT-CAD}, a geometry-aware framework for parametric CAD reconstruction from single- and multi-view observations. We transfer pretrained 3D geometric priors into CAD reconstruction by encoding camera parameters as condition tokens and jointly modeling them with image tokens. To handle varying numbers of viewpoints, we introduce a variable-view cross-view context aggregation module that adaptively fuses multi-view features. We further develop a training-free geometry-aware view selection strategy to select complementary and reliable frames during inference. The resulting representation is decoded into CAD command sequences using a non-autoregressive decoder. We also develop \textbf{VideoCAD}, a large-scale multi-view video benchmark derived from existing CAD data through multi-view re-rendering. Extensive experiments demonstrate the effectiveness of VGGT-CAD for visual CAD reconstruction under different observation configurations.
\end{abstract}


\section{Introduction}
\label{sec:introduction}
3D modeling holds extensive and profound value across multiple industries, including product design~\cite{DBLP:journals/tii/WangWWXX24,DBLP:journals/tii/WangRW26}, architectural engineering~\cite{DBLP:journals/tvcg/CaoSMYXSZLQ26}, animation production~\cite{DBLP:journals/tvcg/HuangTCZCBLL26}, and immersive virtual environments~\cite{DBLP:journals/tvcg/ZangHCWXYZJXC25, DBLP:conf/cvpr/ChenDZYZLPS24}. Traditionally, creating high-quality 3D models has been the exclusive domain of skilled professionals, requiring not only a deep understanding of complex software but also a steep learning curve. The emergence of AI generative model have democratized 3D content creation for a broader audience~\cite{DBLP:conf/cvpr/Lin0TTZHKF0L23}. Among numerous 3D representation methods, parametric Computer-Aided Design (CAD) models occupy an important position in industrial manufacturing, mechanical engineering, and product prototyping. Unlike discrete representations such as polygon meshes or point clouds, CAD models are defined by a sequence of structured and editable geometric operations, such as sketching and extrusion~\cite{DBLP:conf/iccv/WuXZ21, DBLP:journals/tvcg/ZangYZLZZDXC26}. This representation endows CAD models with infinite resolution, precise geometric constraints, and native compatibility with downstream engineering workflows. Accordingly, the automatic generation or reconstruction of parametric CAD models from visual observations has emerged as a critical task, offering a pathway to bridge the gap between visual perception and digital manufacturing pipelines~\cite{DBLP:conf/cvpr/LiLLLZWLWG25}.

\begin{figure}[t]
\vspace{-2mm}
\centering
\includegraphics[width=0.5\textwidth]{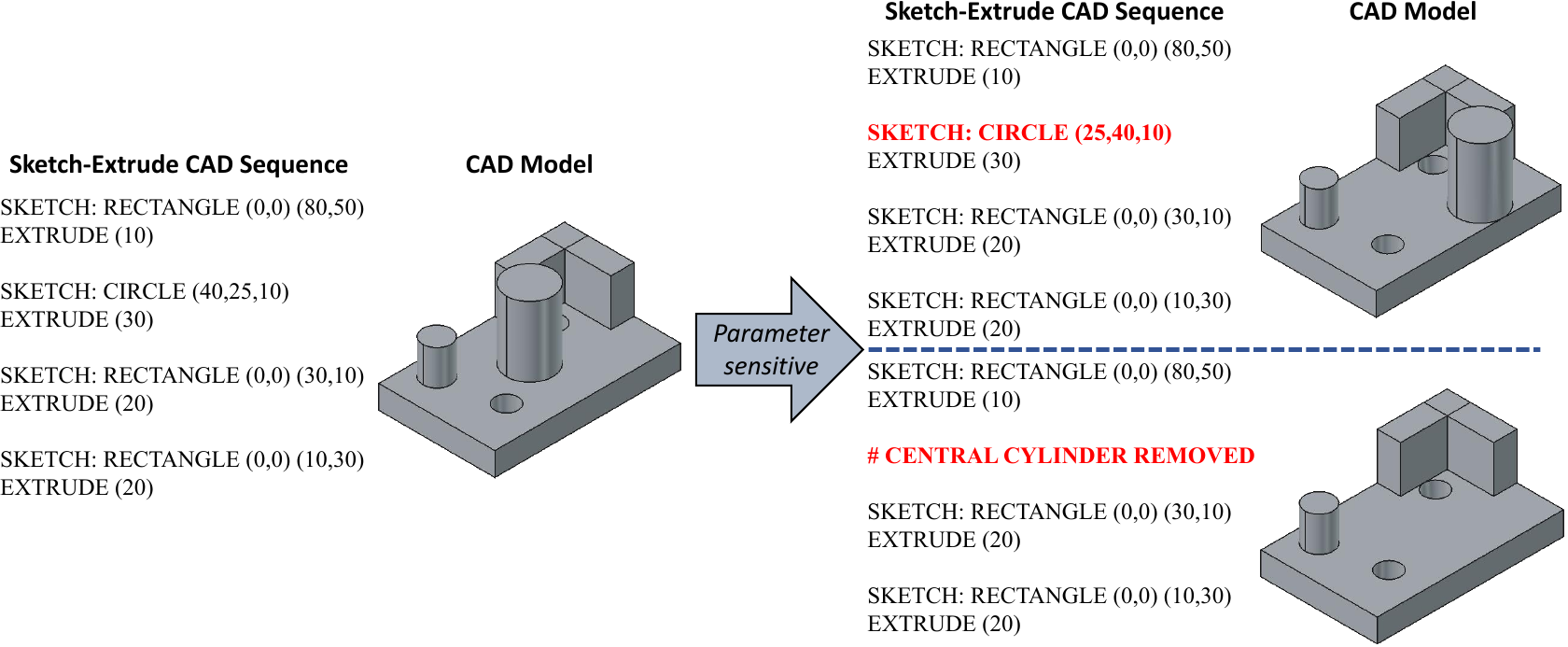}
\caption{\textbf{Program sensitivity analysis.} Parameter changes or removal in CAD commands cause significant geometric variations, highlighting the sensitivity of CAD sequence generation.}
\label{Fig:intro}
\vspace{-6mm}
\end{figure}

However, CAD reconstruction remains difficult because it requires more than predicting a plausible 3D shape. The model must infer an accurate and valid sequence of geometric operations with correct structure, topology, and parameters, as shown in Fig.~\ref{Fig:intro}. Although progress have been made in this field, there is still a huge gap between existing methods and practical systems. We notice that the central issue for existing CAD generation or reconstruction pipeline is the lack of strong \emph{geometric priors}. Existing methods have shown that CAD programs can be modeled effectively as token sequences~\cite{DBLP:conf/iccv/WuXZ21,DBLP:conf/icml/XuWLCJF22}, and recent approaches further enable reconstruction from a single image~\cite{DBLP:journals/tii/ChenYHLXCZZZLS25,DBLP:journals/tmlr/AlamA25,DBLP:conf/cvpr/ChenWC0YZYFLHL25}. However, most of them rely primarily on appearance cues and only implicitly capture 3D structure. For CAD reconstruction, where precise geometry is essential, this is often insufficient.

A natural way to reduce ambiguity is to incorporate multiple views. In the broader 3D vision literature, camera-conditioned multi-view reasoning has proven highly effective for learning geometrically consistent representations~\cite{DBLP:conf/nips/LiuXJCTXS23,DBLP:conf/iclr/ShiWYMLY24,DBLP:conf/cvpr/Wang0CCR24}. More recently, VGGT~\cite{DBLP:conf/cvpr/WangCKV0N25,DBLP:journals/corr/abs-2605-15195} demonstrated strong capability in modeling cross-view geometry and 3D spatial relationships. These advances suggest an important opportunity for CAD reconstruction: instead of treating single-view prediction as the only setting, we should build a unified framework that benefits from learned geometric priors under single-view input and further exploits explicit spatial constraints when multiple views are available.

Motivated by this insight, we propose \textbf{VGGT-CAD}, a unified end-to-end framework for reconstructing \emph{parametric and editable CAD models from either a single image or multiple views}. Our key idea is to transfer geometry-aware priors from large-scale 3D visual understanding into CAD reconstruction through camera-conditioned visual modeling. Specifically, we encode camera intrinsics and extrinsics as condition tokens and fuse them with image tokens in a geometric multi-view visual encoder, enabling the model to reason jointly about appearance, spatial layout, and cross-view consistency. When only a single image is provided, the same framework still benefits from the geometric prior learned by the encoder. When multiple views are available, camera conditioning and cross-view interaction provide stronger structural constraints and substantially reduce reconstruction ambiguity.

Our framework integrates geometry-aware view selection, geometric multi-view visual encoding, cross-view feature aggregation, and CAD sequence prediction into a unified pipeline. For multi-view input, we first select informative views with good geometric coverage and low redundancy with our proposed view selection strategy. A dedicated camera condition encoder then maps camera parameters into high-dimensional tokens, which are jointly processed with image tokens by a VGGT-based Vision Transformer. To support a variable number of input views, we introduce a cross-view context aggregator that fuses per-view features into a compact global representation. Finally, a non-autoregressive Transformer decoder predicts discrete CAD commands and quantized parameters, which can be directly executed by a standard CAD kernel. We further introduce an auxiliary camera reconstruction objective to enhance the encoder's spatial awareness during training.

To evaluate our method, we construct \textbf{VideoCAD}, a new benchmark built upon the CAD corpus of ABC-mono~\cite{DBLP:journals/tii/ChenYHLXCZZZLS25}, whose models are originally sourced from DeepCAD~\cite{DBLP:conf/iccv/WuXZ21} and self-collected data. For each CAD model, we generate 36 views with randomly initialized camera orientations and continuously varying elevation and azimuth angles, while simultaneously recording the corresponding camera parameters. VideoCAD supports systematic evaluation from single-view to multi-view settings. Extensive experiments show that VGGT-CAD consistently outperforms strong baselines and prior single-view methods in both geometric fidelity and CAD command accuracy, demonstrating the effectiveness of explicit geometric priors and camera-aware reasoning for robust parametric CAD reconstruction.

In summary, the main contributions of this paper are as follows:
\begin{itemize}
    \item We develop a geometry-aware CAD reconstruction framework that transfers pretrained 3D visual priors to parametric CAD sequence prediction through explicit camera-conditioned feature encoding, opening up a new research direction for visual CAD reconstruction.

    \item We introduce a variable-view cross-view context aggregation module that adaptively fuses geometric features from a varying number of input viewpoints into a compact global representation, enabling robust CAD sequence generation under different multi-view observation configurations.
    
    \item We propose a training-free geometry-aware view selection strategy, which models input frame selection during the inference stage as a subset optimization problem that balances camera coverage, image reliability, and viewpoint redundancy. This mitigates the reconstruction instability caused by random view sampling and provides more reliable multi-view geometric evidence for CAD sequence generation.
    
    \item We develop VideoCAD, a large-scale multi-view video benchmark built upon existing CAD datasets containing CAD models, multi-view video sequences, and camera annotations, and demonstrate through comprehensive experiments that our method establishes new state-of-the-art performance on the visual CAD reconstruction task.
\end{itemize}

\section{Related works}
\label{sec:related_works}

\subsection{Multi-View 3D Reconstruction and Generation}
    Reviewing general multi-view 3D reconstruction techniques is essential. Traditional algorithms, such as Structure-from-Motion(SfM)~\cite{DBLP:conf/cvpr/SchonbergerF16} and Multi-View Stereo(MVS)~\cite{DBLP:conf/eccv/SchonbergerZFP16}, reconstruct dense point clouds via multi-view geometric matching. Recently, implicit neural rendering paradigms, notably NeRF~\cite{DBLP:conf/eccv/MildenhallSTBRN20} and 3D Gaussian Splatting (3DGS)~\cite{DBLP:journals/tog/KerblKLD23}, have revolutionized novel-view synthesis and high-fidelity reconstruction. However, both traditional MVS and implicit radiance fields typically produce uneditable dense meshes or continuous implicit representations. These outputs lack the sharp geometric edges and parametric editability critical for industrial manufacturing, rendering them unsuitable for direct integration into downstream CAD software.
    
    Alternatively, feed-forward 3D large models based on diffusion models or Transformers have recently emerged. Frameworks like LRM~\cite{DBLP:conf/iclr/Hong0GBZLLSB024} and InstantMesh~\cite{DBLP:journals/corr/abs-2404-07191} swiftly predict triplane representations and extract meshes from sparse views via a single forward pass. Furthermore, recent large-scale 3D synthesis systems, such as Hunyuan3D 2.0~\cite{DBLP:journals/corr/abs-2501-12202}, scale a flow-based diffusion Transformer alongside potent geometric and diffusion priors to generate high-resolution, textured 3D assets. Nevertheless, because these architectures lack explicit multi-view camera conditioning and keyframe filtering mechanisms for sequential video redundancy, they still struggle to robustly infer accurate underlying topologies when confronted with real-world video sequences featuring severe self-occlusion.

\subsection{Generative Models for CAD}
    Recently, deep learning-based generation and reconstruction of Computer-Aided Design (CAD) models have witnessed significant progress. Based on the evolution of input modalities and generative paradigms, existing research can be broadly categorized into several key developmental stages, transitioning from unconditional generation to multimodal conditional constraints.
    
    \subsubsection{From Unconditional Generation to 3D Prior-Driven Modeling} An early milestone in this field was formulating the complex CAD modeling process as the generation of structured, discrete command sequences. DeepCAD~\cite{DBLP:conf/iccv/WuXZ21} pioneered a Transformer-based autoencoder to cast 3D shape generation as a sequence modeling task for CAD commands. Subsequently, SkexGen~\cite{DBLP:conf/icml/XuWLCJF22} introduced an autoregressive paradigm, while HNC-CAD~\cite{DBLP:conf/icml/XuJLWF23} proposed a three-tier hierarchical neural encoding to enhance local geometric control. However, these methods primarily focus on random synthesis and cannot reverse-engineer specific real-world objects. To achieve precise instance-level reconstruction, later studies turned to complete 3D data. Methods like ExtrudeNet~\cite{DBLP:conf/eccv/RenZCLZ22}, Point2CAD~\cite{DBLP:conf/cvpr/LiuOWS24}, and CAD-SIGNet~\cite{DBLP:conf/cvpr/KhanDAC0A24} achieve accurate mappings from 3D inputs to parametric commands. Despite their high precision, these approaches rely heavily on expensive 3D scanning equipment to acquire high-quality point clouds, posing a significant barrier to broad, everyday application.
    
    \subsubsection{Image Conditioned Generation} To lower the barrier of physical data acquisition, recent works have explored generating CAD models solely from 2D images. Methods such as Img2CAD~\cite{DBLP:journals/tii/ChenYHLXCZZZLS25} and GenCAD~\cite{DBLP:journals/tmlr/AlamA25} generate parametric CAD models from single images, while CADCrafter~\cite{DBLP:conf/cvpr/ChenWC0YZYFLHL25} further explores multi-view image conditions with diffusion-based generation, significantly enhancing user interaction. However, these methods primarily rely on incomplete 2D appearance cues with limited 3D geometric priors, making the recovery of occluded geometry challenging, particularly when critical regions are not sufficiently observed.
    %However, these advances often rely excessively on 2D appearance cues and lack robust 3D geometric priors. Consequently, when confronted with single-view blind spots such as self-occlusion, these models inevitably suffer from severe geometric ambiguity, hindering stable and high-fidelity topological reconstruction.
    
    \subsubsection{LLM-Driven CAD Modeling} As Large Language Models(LLMs) demonstrate exceptional capabilities in code generation and logical reasoning, CAD modeling has entered a new dimension. Researchers have begun leveraging LLMs to process highly structured CAD languages. CAD-LLM~\cite{DBLP:conf/nips/WuKKJPWL23} first demonstrated that fine-tuned natural language models could process engineering sketches, while Text2CAD~\cite{DBLP:conf/nips/KhanSSSAA24} introduced the first framework translating natural language prompts into parametric CAD sequences. Recently, this direction has seen breakthrough progress; FlexCAD~\cite{DBLP:conf/iclr/ZhangS00B25} represents CAD models as structured text, enabling controllable generation across all construction levels via mask-based fine-tuning; CAD-Llama~\cite{DBLP:conf/cvpr/LiMLLZZ25} introduced the Structured Parametric Code(SPCC) format, endowing LLMs with preliminary spatial awareness through instruction tuning. Furthermore, CADFusion~\cite{DBLP:conf/icml/WangYS025} innovatively integrated visual feedback from rendered images into the LLM training loop, and CADCodeVerify~\cite{DBLP:conf/iclr/AlrashedyTZLXG25} utilized Vision-Language Models(VLMs) to iteratively verify and refine generated CAD scripts. These works underscore the immense potential of LLMs in complex CAD sequence generation.
    
    Despite these advancements, existing conditional generation methods have not yet resolved a core challenge: \textbf{How to endow networks with robust 3D spatial understanding and effectively leverage multi-view observations to provide rigorous structural constraints.} This critical aspect remains underexplored in the current literature.

\begin{figure*}[htb]
\centering
\includegraphics[width=0.95\textwidth]{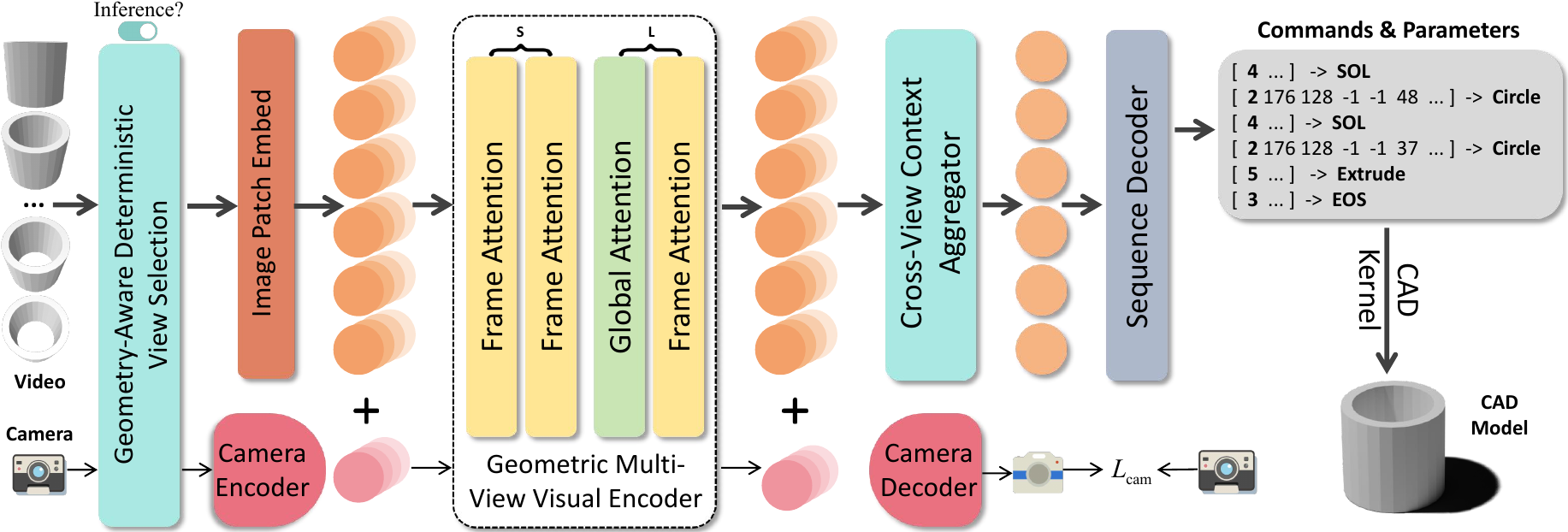}
\vspace{-1.5mm}
\caption{\textbf{VGGT-CAD framework.} The framework consists of five components: (1) Geometry-Aware Deterministic View Selection for selecting informative and complementary views; (2) Camera Condition Encoder and Geometric Multi-View Visual Encoder for camera-aware multi-view feature extraction; (3) Variable-View Cross-View Context Aggregator for adaptively fusing features from different numbers of views into a global latent representation $z$; (4) Transformer-based Sequence Decoder for non-autoregressive CAD command prediction; and (5) an auxiliary Camera Condition Decoder for camera pose reconstruction supervision during training.
}
\label{Fig:piplint}
\vspace{-6mm}
\end{figure*}

\section{Method}
\label{sec:method}
    We formulate the reconstruction problem in Section~\ref{subsec:formulation}. In Section~\ref{subsec:command_sequencec_representation}, we define the CAD command sequence representation used by the model. Section~\ref{subsec:Network_Architecture} presents the network architecture of VGGT-CAD, including camera condition encoding, the geometric multi-view visual encoder, the cross-view context aggregator, and the sequence decoder. We then describe the training strategy and optimization objectives in Section~\ref{subsec:optimization_training}. Finally, Section~\ref{subsec:GAD_view_selection} introduces the geometry-aware deterministic view selection used to choose informative input views at inference time.

\subsection{Formulation}
\label{subsec:formulation}
Given a monocular video $V$ capturing a target object, our goal is to generate a command sequence $M$ that describes the 3D geometric structure of the object. To extract effective spatial geometric information from the continuous video stream, we first sample the video $V$, dividing it into an image sequence of $T = 36$ frames. In each training iteration, we randomly sample $N$ images from these $T$ frames to form a multi-view input set $\mathcal{I} = \{I_1, I_2, \dots, I_N\}$, where $N \in [2, 8]$, and each image $I_i \in \mathbb{R}^{H \times W \times 3}$.

In addition to the images themselves, previous studies have shown that camera viewpoint information can provide a strong spatial prior for the model, helping it more accurately understand the relative relationships between the global structure and local features of the object. Therefore, we also use the extrinsic and intrinsic parameters of the corresponding camera pose for each viewpoint as input conditions, denoted as $P = \{P_1, P_2, \dots, P_N\}$.

Our generation target $M$ is an operation sequence of length $N_c$, i.e., $M = [C_1, C_2, \dots, C_{N_c}]$. Each operation command $C_j$ contains a discrete command type $t_j$ and a corresponding continuous or discrete parameter vector $p_j$. Ultimately, the generated command sequence $M$ is executed by a standard kernel parser to render the final 3D solid model.

The entire VGGT-CAD framework can be formulated as a conditional probability generative model:
\begin{align}
    \begin{split}
    p(M|\mathcal{I},P,\Theta) = \prod_{j=1}^{N_c} p(t_j, p_j|\mathcal{I}, P, \Theta)
    \end{split}
\end{align}
where $\Theta$ represents all learnable parameters in the network.

\subsection{Command Sequence Representation}
\label{subsec:command_sequencec_representation}

To enable neural networks to effectively learn and generate geometric models, we transform complex boundary representations into command sequences akin to natural language. We focus on the two most core operations in industrial design, namely sketch and extrude. A complete construction process is defined as an ordered set of a series of operations. Specifically, each command $C_j$ is represented as a tuple $(t_j, p_j)$.

The command type $t_j$ encompasses structure control commands, 2D curve drawing commands, and 3D generation commands. To unify the parameter formats across different commands, We stack all possible parameters into a fixed-length 16-dimensional vector $p_j = [x, y, \alpha, f, r, \theta, \phi, \gamma, p_x, p_y, p_z, s, e_1, e_2, b, u]$. These parameters cover 2D coordinates, angles, extrusion distances, boolean operation types, and so on, as shown in Tab.~\ref{tab:data_rep_new1}. For parameter slots unused by a specific command, we uniformly pad them with a specific invalid value of $-1$.

To avoid the disruption of geometric constraints caused by directly regressing continuous parameters, we normalize the bounding box of the entire model into the $[-1, 1]^3$ space and quantize all continuous parameters into 256 discrete levels. This quantization strategy transforms the complex continuous parameter regression problem into a discrete classification problem that is much easier for network optimization.

\begin{table}[h]
\centering
\caption{\textbf{CAD Command Dictionary and Parameter Definitions.} Commands are categorized into control tokens, 2D sketch primitives, and 3D extrusions. $\cmdSOL$ denotes the start of a loop, and $\cmdEOS$ terminates the sequence.}
\label{tab:data_rep_new1}
\resizebox{\columnwidth}{!}{%
\begin{tabular}{@{}llcl@{}}
\toprule
\textbf{Category} & \textbf{Command} & \textbf{Param.} & \textbf{Description} \\ \midrule
\multirow{2}{*}{\textit{Control}} & $\cmdSOL$ & -- & Start of a loop \\
 & $\cmdEOS$ & -- & End of the whole sequence \\ \midrule
\multirow{6}{*}{\textit{2D Sketch}} & $\cmdLine$ & $x, y$ & End-point coordinates \\ \cmidrule(l){2-4} 
 & \multirow{3}{*}{$\cmdArc$} & $x, y$ & Arc end-point coordinates \\
 &  & $\alpha$ & Sweep angle \\
 &  & $f$ & Counter-clockwise flag \\ \cmidrule(l){2-4} 
 & \multirow{2}{*}{$\cmdCirc$} & $x, y$ & Center coordinates \\
 &  & $r$ & Radius \\ \midrule
\multirow{5}{*}{\textit{3D Extrude}} & \multirow{5}{*}{$\cmdExt$} & $\theta, \phi, \gamma$ & Sketch plane orientation \\
 &  & $p_x, p_y, p_z$ & Sketch plane origin \\
 &  & $s$ & Scale of associated sketch profile \\
 &  & $e_1, e_2$ & Extrude distances toward both sides \\
 &  & $b, u$ & Boolean type ($b$) and extrude type ($u$) \\ \bottomrule
\end{tabular}%
}
\vspace{-8mm}
\end{table}

\subsection{Network Architecture}
\label{subsec:Network_Architecture}
The VGGT-CAD architecture consists of five learnable modules, as illustrated in Fig.~\ref{Fig:piplint}. First, the camera condition encoder converts the camera parameters associated with each selected viewpoint into explicit camera-conditioned spatial tokens. The geometric multi-view visual encoder then integrates these tokens with visual features to capture geometric relationships both within individual views and across views. The cross-view context aggregator consolidates the resulting viewpoint and camera tokens into a fixed-dimensional global geometric representation while accommodating a variable number of input viewpoints. In parallel, the camera condition decoder reconstructs the camera parameters from the corresponding camera features to provide explicit geometric supervision. Finally, a Transformer-based non-autoregressive sequence decoder maps the global geometric representation to the CAD construction sequence by predicting command tokens at predefined positions in parallel.

\subsubsection{Camera Condition Encoding}
\label{subsubsec:camera_condition_encoding}
In multi-view 3D reconstruction and generation tasks, camera tokens have been proven to effectively provide relative geometric relationships between viewpoints, serving as a key factor in enhancing the model's spatial awareness. Therefore, we design a dedicated encoder to process the input camera pose parameters $P$.

For the $i$-th viewpoint, given its camera extrinsic and intrinsic matrices, we first compute its inverse affine transform to obtain the camera-to-world pose encoding, and combine it with the intrinsic information and image dimensions to generate an initial pose feature vector. Subsequently, this feature vector is dimensionally mapped through a multi-layer perceptron (MLP) branch and then fed into a Transformer trunk with a depth of $L_{\text{cam}}$ for iterative refinement. After Layer Normalization, the high-dimensional camera token $c_i \in \mathbb{R}^d$ is finally output. This process can be expressed as:
\begin{align}
    \begin{split}
c_i = \text{Trunk}(\text{LayerNorm}(\text{MLP}(\text{PoseEncoding}(P_i))))
    \end{split}
\end{align}
In practical applications, camera pose parameters can be estimated in real time during image capture using mobile AR frameworks such as ARCore, ARKit, and AREngine.
\footnote{\textit{ARCore: \url{https://developers.google.com/ar}}

\textit{ARKit: \url{https://developer.apple.com/augmented-reality/arkit/}}

\textit{AREngine:\url{https://developer.huawei.com/consumer/en/hms/huawei-arengine/}}}

% \subsubsection{Geometric Visual Encoder Enabling Multi-View Input}
\subsubsection{Geometric Multi-View Visual Encoder}
\label{subsubsec:multi-view_visual_encoder}
CAD models are essentially engineering representations composed of precise geometric constraints, topological relationships, and editable operation sequences, rather than merely 3D surfaces that require visual appearance similarity. Therefore, the model needs to understand the structural consistency of an object across different viewpoints. We adopt the multi-view visual encoding structure from VGGT as our core visual backbone, introducing the geometric prior learned from its multi-view 3D understanding into the parametric CAD reconstruction task. This geometric prior enables the model to model cross-view consistency, camera geometric relationships, and 3D spatial correspondences within the feature space, thereby no longer relying solely on 2D appearance correlations for shape inference.

Given an input image $I_i$, we partition it into non-overlapping patches and map them into an image token sequence $X_i = [x_{i,1}, x_{i,2}, \dots, x_{i,M}] \in \mathbb{R}^{M \times d}$, where $M$ is the number of patches and $d$ is the token dimension. To incorporate the camera token $c_i \in \mathbb{R}^d$ obtained in Sec.~\ref{subsubsec:camera_condition_encoding} into the encoding process, we concatenate it with the image tokens along the sequence dimension to form the complete input sequence for this viewpoint:
\begin{align}
    \begin{split}
S_i = [c_i \parallel x_{i,1}, x_{i,2}, \dots, x_{i,M}] \in \mathbb{R}^{(1+M) \times d}
    \end{split}
\end{align}
where $\parallel$ denotes the concatenation operation along the sequence dimension.

In the backbone network, these camera-informative token sequences $S_1, S_2, \dots, S_N$ not only undergo within-view self-attention computations within each individual viewpoint but also engage in cross-view self-attention interactions at specific transformer layers. For a token $x_k^{(i)}$ in viewpoint $i$, the cross-view self-attention computation can be expressed as:
\begin{align}
    \begin{split}
\text{Attention}(x_k^{(i)}, X) = \text{Softmax}\left(\frac{Q_k^{(i)} K^T}{\sqrt{d}}\right) V
    \end{split}
\end{align}
where $X = \{S_1, S_2, \dots, S_N\}$ is the set containing the complete token sequences of all viewpoints, $Q_k^{(i)}$ is the query vector, and $K$ and $V$ are the key and value matrices generated from the tokens of all viewpoints. Since the camera token $c_i$ participates in the attention computation alongside the image tokens, viewpoint orientation information is naturally propagated to all image features during this process, allowing the network to implicitly establish geometric correspondences across different views. Furthermore, thanks to the geometric inductive bias and camera-conditioned constraints provided by VGGT, the model can more accurately fuse multi-view evidence, alleviating common issues in single-view methods such as back-facing structural hallucinations, missing topology, and unstable parameter predictions.

After multi-layer processing by the encoder, we obtain the output sequence $S_i \in \mathbb{R}^{(1+M) \times d}$ for each viewpoint. We split it into camera features and image features based on the position indices during concatenation:
\begin{equation}
\begin{aligned}
c_i &= S_i[0] \in \mathbb{R}^{d}, \\
F_i &= S_i[1:] \in \mathbb{R}^{M \times d},
\end{aligned}
\end{equation}
Since the ViT divides the original image into $h \times w$ patches, yielding a total of $M=h \times w$ tokens, we reshape the one-dimensional image token sequence $F_i$ into a 2D spatial feature map to preserve spatial structural information:
\begin{align}
    \begin{split}
F_i = \text{Reshape}(F_i) \in \mathbb{R}^{h \times w \times d}
    \end{split}
\end{align}
The set of image features $\mathcal{F}_{\text{img}} = {F_1, F_2, \dots, F_N}$ is first fed into the Cross-View Context Aggregator in Sec.~\ref{subsubsec:cross-view_context_aggregator}, whose output is then passed to the Sequence Decoder in Sec.~\ref{subsubsec:sequence_decoder}. Meanwhile, the isolated set of camera features $\mathcal{F}_{\text{cam}} = \{c_1, c_2, \dots, c_N\}$ is fed into the camera condition decoder in Sec.~\ref{subsubsec:camera_condition_decoding} to construct auxiliary pose reconstruction supervision signals.

\subsubsection{Camera Condition Decoding}
\label{subsubsec:camera_condition_decoding}
To further enhance the visual encoder's ability to understand spatial poses, we designed a dedicated decoder to decode the camera features $\mathcal{F}_{\text{cam}}$ separated in Sec.~\ref{subsubsec:multi-view_visual_encoder} and compare them with the ground-truth camera parameters, thereby constructing an auxiliary pose reconstruction supervision signal.

This decoder uses a two-layer fully connected network as its backbone, followed by three independent prediction heads to regress different camera parameter components. For the camera feature $c_i \in \mathbb{R}^d$ of the $i$-th viewpoint, the decoding process can be expressed as:

\begin{equation}
\begin{aligned}
t_i &= W_t g(c_i) + b_t, & t_i &\in \mathbb{R}^{3},\\
q_i &= W_q g(c_i) + b_q, & q_i &\in \mathbb{R}^{4},\\
f_i^{\mathrm{FOV}} &= \operatorname{ReLU}\!\left(W_f g(c_i) + b_f\right), & f_i^{\mathrm{FOV}} &\in \mathbb{R}^{2}.
\end{aligned}
\end{equation}
where $g(\cdot)$ represents the two-layer fully connected backbone network with ReLU activation, $t_i$ is the predicted translation vector, $q_i$ is the predicted rotation quaternion, and $f_i^{\mathrm{FOV}}=[\mathrm{FOV}_{w,i},\mathrm{FOV}_{h,i}]$ denotes the predicted horizontal and vertical fields of view. The FOV prediction head uses a ReLU activation to ensure non-negative outputs. The predicted FOVs are subsequently converted into pixel-unit focal lengths according to
\begin{equation}
\begin{aligned}
\hat{f}_{x,i}=\frac{W}{2\tan(\mathrm{FOV}_{w,i}/2)},\\
\hat{f}_{y,i}=\frac{H}{2\tan(\mathrm{FOV}_{h,i}/2)}.
\end{aligned}
\end{equation}
The resulting focal lengths are used to form the intrinsic camera parameters and are supervised in the focal-length space. Thus, the complete camera pose encoding is represented as $P_i=[t_i|q_i|f_i^{\mathrm{FOV}}]\in\mathbb{R}^{9}$, while the camera reconstruction loss is computed from the corresponding camera parameters, as detailed in Sec.~\ref{subsec:optimization_training}.

\subsubsection{Cross-View Context Aggregator}
\label{subsubsec:cross-view_context_aggregator}
Since the number of input viewpoints $N$ dynamically varies between 2 and 8, and the subsequent CAD sequence decoder requires a fixed-dimensional global context representation, we designed a Cross-View Context Aggregator to replace simple feature concatenation.

This module receives the multi-view image features $\mathcal{F}_{\text{img}}$ from the geometric multi-view visual encoder. First, we use a shared-weight 2D convolutional encoder to process the features $F_i$ of each viewpoint independently. This convolutional encoder consists of multiple convolution layers, batch normalization, and ReLU activation functions, and uses adaptive average pooling at the end to compress the spatial dimensions, ultimately flattening them to obtain a 1D feature vector $v_i \in \mathbb{R}^{d_v}$ for each viewpoint.

To adaptively fuse the features of different viewpoints, we introduce an attention network based on a MLP. This network calculates the attention score for each viewpoint feature and normalizes it across the viewpoint dimension using a softmax function, ensuring that the sum of the weights for all viewpoints equals 1:
\begin{equation}
\begin{aligned}
\alpha_i = \frac{\exp(\text{MLP}_{\text{attn}}(v_i))}{\sum_{j=1}^N \exp(\text{MLP}_{\text{attn}}(v_j))}
\end{aligned}
\end{equation}

Subsequently, we use these attention weights to perform a weighted sum on the multi-view features to obtain the fused feature:
\begin{equation}
\begin{aligned}
f_{\text{fused}} = \sum_{i=1}^N \alpha_i v_i
\end{aligned}
\end{equation}

Finally, the fused feature $f_{\text{fused}}$ is passed through a linear projection layer for dimensionality reduction, outputting the final global latent vector $z \in \mathbb{R}^{d_z}$:
\begin{equation}
\begin{aligned}
z = W_{\text{proj}} f_{\text{fused}} + b_{\text{proj}}
\end{aligned}
\end{equation}

This vector $z$ compactly encodes the complete 3D geometric structural prior of the target object within the input video clip.

\subsubsection{Sequence Decoder}
\label{subsubsec:sequence_decoder}
After obtaining the global latent vector $z$, we employ a Transformer-based decoder to generate the command sequence. The decoder adopts a feed-forward non-autoregressive strategy. It receives a set of learnable constant embeddings $E \in \mathbb{R}^{N_c \times d_E}$ of length $N_c$ as input queries, and interacts with the global latent vector $z$ through a cross-attention mechanism. The update process for the $l$-th layer of the decoder is:
\begin{equation}
\begin{aligned}
H^{(l)} = \text{FFN}(\text{CrossAttn}(\text{SelfAttn}(H^{(l-1)}), z))
\end{aligned}
\end{equation}
where $H^{(0)} = E$. The final output of the decoder $H^{(L)}$ passes through two parallel linear prediction heads. The command classification head predicts the command type distribution $P(t_j) = \text{Softmax}(W_t H_j^{(L)} + b_t)$ for each time step $j$. The parameter classification head predicts the 16 parameters of the corresponding command. Since the parameters have been quantized into 256 levels, the output dimension of this prediction head is $16 \times 256$, meaning it independently predicts the probability distribution over 256 discrete categories for each parameter:

\subsection{Optimization and Training}
\label{subsec:optimization_training}
During the training phase, we adopt an end-to-end supervised learning strategy. The total loss function $\mathcal{L}_{\text{total}}=\mathcal{L}_{\text{CAD}}+\lambda\mathcal{L}_{\text{cam}}$, where set $\lambda=1.0$. Since the parameters have been quantized into discrete categories, we treat the predictions of both command types and command parameters as multi-class classification tasks. The CAD sequence generation loss consists of the cross-entropy losses for command types and command parameters:
\begin{equation}
\begin{aligned}
\mathcal{L}_{\text{type}}=\frac{1}{N_c}\sum_{j=1}^{N_c}\mathcal{H}(\hat{t}_j,t_j)
\end{aligned}
\end{equation}
\begin{equation}
\begin{aligned}
\mathcal{L}_{\text{param}}=\frac{1}{N_c}\sum_{j=1}^{N_c}\frac{1}{|K_j|}\sum_{k\in K_j}\mathcal{H}(\hat{p}_{j,k},p_{j,k})
\end{aligned}
\end{equation}
\begin{equation}
\begin{aligned}
\mathcal{L}_{\text{CAD}}=\mathcal{L}_{\text{type}}+\beta\mathcal{L}_{\text{param}}
\end{aligned}
\end{equation}
where $\mathcal{H}(\cdot,\cdot)$ denotes the cross-entropy function, $\hat{t}_j$ is the predicted command type distribution at the $j$-th time step, and $t_j$ is the corresponding ground-truth label. $\hat{p}_{j,k}$ is the predicted probability distribution over the 256 discrete categories for the $k$-th parameter of the $j$-th command, and $p_{j,k}$ is the corresponding ground-truth category label. $K_j$ represents the set of actual valid parameter indices for the $j$-th command, and $|K_j|$ is its cardinality. For parameter slots unused by a specific command type as well as padding positions at the end of the sequence, we exclude them from the loss calculation via a Validity Mask, avoiding the interference of invalid supervision signals on network optimization. $\beta=2$ is a balancing coefficient used to adjust the relative importance between type classification and parameter classification.

To encourage the geometric multi-view visual encoder to better understand spatial poses, we contrast the prediction results $P_i=[\hat{t}_i|\hat{q}_i|\hat{f}_i]$ of the camera decoder from Sec.~\ref{subsubsec:camera_condition_decoding} with the ground-truth camera parameters. The camera loss $\mathcal{L}_{\text{cam}}$ consists of three components: translation loss $\mathcal{L}_t$, rotation loss $\mathcal{L}_q$, and focal length loss $\mathcal{L}_f$:
\begin{equation}
\begin{aligned}
\mathcal{L}_t=\frac{1}{N}\sum_{i=1}^N\sum_{d=1}^3|\hat{t}_{i,d}-t_{i,d}|
\end{aligned}
\end{equation}
\begin{equation}
\begin{aligned}
\mathcal{L}_q=\frac{1}{N}\sum_{i=1}^N\min\left(\sum_{d=1}^4|\hat{q}_{i,d}-q_{i,d}|,\sum_{d=1}^4|\hat{q}_{i,d}+q_{i,d}|\right)
\end{aligned}
\end{equation}
\begin{equation}
\begin{aligned}
\mathcal{L}_f=\frac{1}{N}\sum_{i=1}^N\left(\left|\frac{\hat{f}_{x,i}}{W}-\frac{f_{x,i}}{W}\right|+\left|\frac{\hat{f}_{y,i}}{H}-\frac{f_{y,i}}{H}\right|\right)
\end{aligned}
\end{equation}
\begin{equation}
\begin{aligned}
\mathcal{L}_{\text{cam}}=\mathcal{L}_t+\mathcal{L}_q+\mathcal{L}_f
\end{aligned}
\end{equation}
where $\hat{t}_i$ and $t_i$ are the predicted and ground-truth translation vectors, respectively; $\hat{q}_i$ and $q_i$ are the predicted and ground-truth rotation quaternions. The $\min$ operation in the rotation loss accounts for the Double Cover property of quaternions, meaning $q$ and $-q$ represent the same rotation. $\hat{f}_{x,i},\hat{f}_{y,i}$ and $f_{x,i},f_{y,i}$ are the predicted and ground-truth focal lengths in the horizontal and vertical directions, respectively. $W$ and $H$ are the image width and height; dividing by the image dimensions normalizes the pixel-unit focal lengths into dimensionless relative values, ensuring scale consistency with the translation and rotation components.

\textbf{Training.} Since the geometric multi-view visual encoder has already learned strong 3D geometric priors from large-scale pretraining, training it from scratch or fully fine-tuning all parameters may damage the pretrained weights and cause catastrophic forgetting. Therefore, we use LoRA to fine-tune only the backbone of the geometric multi-view visual encoder in VGGT-CAD, which reduces VRAM usage and preserves its generalization ability. In contrast, the camera encoder and decoder, cross-view context aggregator, and sequence decoder are trained from scratch to better adapt to the command generation task. 

\subsection{Geometry-Aware Deterministic View Selection}
\label{subsec:GAD_view_selection}
Videos usually contain redundant frames, while CAD program recovery only requires a small set of geometrically complementary and visually reliable viewpoints. During training, random frame sampling serves as data augmentation and improves robustness to viewpoint combinations. However, at inference time, random sampling may introduce inconsistent inputs and increase output variance. Therefore, inspired by next-best-view planning~\cite{DBLP:journals/pami/DenzlerB02}, information gain maximization~\cite{DBLP:journals/jmlr/KrauseSG08}, and coverage-based subset selection~\cite{DBLP:journals/mp/NemhauserWF78}, we formulate test-time frame selection as a deterministic observation subset optimization problem under a fixed cardinality budget, aiming to maximize the structural evidence retained from candidate frames.

Given a candidate frame set $\mathcal{V}=\{(I_i, K_i, T_i)\}_{i=1}^T$, where $I_i$, $K_i$, and $T_i$ denote the $i$-th frame image, camera intrinsics, and camera pose, respectively. Ideally, the optimal viewpoint set should maximize its information gain with respect to the latent CAD structure, i.e., selecting $S^\star = \arg\max_{S \subseteq \mathcal{V}} I(S; G)$ s.t. $|S|=K$. Here, $K$ is the viewpoint budget. Since the true CAD structure is invisible during inference, directly computing the mutual information is infeasible. Therefore, we use a computable surrogate marginal utility function to approximate this objective, and incrementally construct the viewpoint set through a deterministic greedy strategy. Specifically, at the $t$-th selection round, for any candidate frame $i \in \mathcal{V} \setminus S_{t-1}$, where $\mathcal{V}$ is the complete set of candidate viewpoints and $S_{t-1}$ is the set of viewpoints already selected after the $(t-1)$-th round, its marginal selection score is defined as:
\begin{equation}
\begin{aligned}
\Delta(i \mid S_{t-1}) = \lambda_c C(i \mid S_{t-1}) + \lambda_q Q(i) - \lambda_r R(i \mid S_{t-1})
\end{aligned}
\end{equation}
where $C$, $Q$, and $R$ represent the geometric coverage gain, image quality, and viewpoint redundancy, respectively, all normalized to $[0,1]$. The final selection process is:
\begin{equation}
\begin{aligned}
i_t &= \arg\max_{i\in\mathcal{V}\setminus S_{t-1}}
\Delta(i\mid S_{t-1}),\\
S_t &= S_{t-1}\cup\{i_t\}, \qquad t=1,\ldots,K.
\end{aligned}
\end{equation}
\textbf{Geometric Coverage Gain.} CAD reconstruction relies on sufficient observation of object surfaces, contours, and self-occluded structures; therefore, the value of a newly added viewpoint depends on whether it supplements geometric evidence not yet covered by the current set. We measure the geometric complementarity between the candidate viewpoint and the selected viewpoints from three aspects: the angle between camera optical axes, the baseline between camera centers, and the horizontal azimuth difference:
\begin{equation}
\begin{aligned}
C(i \mid S) = \lambda_1^c C_{\text{axis}}(i \mid S) + \lambda_2^c C_{\text{center}}(i \mid S) + \lambda_3^c C_{\text{azim}}(i \mid S)
\end{aligned}
\end{equation}
where $C_{\text{axis}}$ measures the difference in camera optical axis directions, $C_{\text{center}}$ measures the effective baseline distance between camera centers, and $C_{\text{azim}}$ measures the coverage complementarity in the horizontal orbit direction. This term encourages the selection of frames that can observe new surfaces, supplement side/back structures, and alleviate single-view occlusion ambiguity.

\textbf{Image Quality Term.} Not all geometrically complementary frames are suitable as CAD decoding evidence. Frames with motion blur, abnormal exposure, or too small of a target proportion will reduce the reliability of cross-view aggregation. Therefore, we use training-free image statistics to estimate single-frame reliability:
\begin{equation}
\begin{aligned}
Q(i) = \lambda_1^q Q_{\text{sharp}}(i) + \lambda_2^q Q_{\text{entropy}}(i) + \lambda_3^q Q_{\text{fg}}(i)
\end{aligned}
\end{equation}
where $Q_{\text{sharp}}$ measures sharpness via Laplacian variance, $Q_{\text{entropy}}$ measures texture and exposure information via brightness entropy, and $Q_{\text{fg}}$ represents the target foreground proportion. This term prioritizes allocating the limited viewpoint budget to frames with more stable visual evidence.

\textbf{Redundancy Penalty Term.} Adjacent frames in continuous videos often have highly similar camera poses; repeatedly selecting such frames increases the input length without significantly adding structural information. To explicitly model diminishing marginal returns, we define a redundancy penalty based on pose similarity:
\begin{equation}
\begin{aligned}
R(i \mid S) = \lambda_1^r R_{\text{axis}}(i \mid S) + \lambda_2^r R_{\text{center}}(i \mid S) + \lambda_3^r R_{\text{azim}}(i \mid S)
\end{aligned}
\end{equation}
where $R_{\text{axis}}$, $R_{\text{center}}$, and $R_{\text{azim}}$ measure the similarity between the candidate viewpoint and the already selected set in terms of optical axis direction, camera center position, and azimuth angle, respectively. This term is complementary to the geometric coverage gain: the former rewards newly added structural evidence, while the latter suppresses repeated observations, thereby avoiding continuous adjacent frames taking up too much budget in the input set. This module does not change the expressive power of the subsequent encoder or CAD decoder, but rather improves their input evidence conditions, ensuring that camera condition encoding, cross-view attention, and CAD sequence generation are built upon more stable structural observations.

\section{Experiments}
\label{sec:experiments}

\subsection{Experimental Setups}
\label{subsec:experimental_setups}
\subsubsection{Datasets}
\label{subsec:Datasets}
Most existing CAD datasets provide parametric models or construction sequences but lack multi-view visual sequences with corresponding camera information, limiting their applicability to video-based CAD reconstruction. To address this limitation, we develop \textit{VideoCAD} based on the existing ABC-mono~\cite{DBLP:journals/tii/ChenYHLXCZZZLS25} CAD corpus, rather than collecting CAD models from scratch. For each CAD model, we generate 36 views with randomly initialized camera orientations and continuously varying elevation and azimuth angles across views, while simultaneously recording the corresponding camera parameters and CAD construction sequence. Following the original data split of ABC-mono, the 208,853 unique CAD models are divided into training, validation, and test sets with a 90\%/5\%/5\% split, respectively, yielding 7,518,708 rendered frames. Each frame is associated with its corresponding camera parameters and CAD construction sequence, providing a unified benchmark for single-view and multi-view video-to-CAD reconstruction.

\subsubsection{Implementation Details}
\label{subsubsec:implementation_details}
Our VGGT-CAD framework is deployed on the PyTorch platform and trained on a computing node equipped with 4 NVIDIA A800 (80GB) GPUs. Model optimization uses the Adam~\cite{DBLP:conf/iclr/LoshchilovH19} optimizer with a batch size of 4. During the training phase, for each CAD model, we randomly sample 2 to 8 frames from the 36 rendered viewpoints and arrange them in chronological order to simulate a casual video input shot by a user. The resolution of the input images is uniformly resized to 512x512. To efficiently adapt the pre-trained visual backbone network, we employ LoRA~\cite{DBLP:conf/iclr/HuSWALWWC22} fine-tuning in the geometric multi-view visual encoder section, with the rank size set to $r=32$. The initial learning rate is set to 0.001.

\subsubsection{Evaluation Metrics}
\label{subsubsec:evaluation_metrics}

Following DeepCAD~\cite{DBLP:conf/iccv/WuXZ21}, we evaluate the generated CAD sequences from both sequence-level and geometry-level perspectives.

At the sequence level, we evaluate Command Accuracy ($ACC_{\text{cmd}}$) and Parameter Accuracy ($ACC_{\text{param}}$). $ACC_{\text{cmd}}$ measures the proportion of correctly predicted command types, while $ACC_{\text{param}}$ evaluates parameter accuracy only for correctly predicted commands:
\begin{equation}
\begin{aligned}
ACC_{\text{cmd}}&=\frac{1}{N_c}\sum_{i=1}^{N_c}\mathbb{I}[\hat{t}_i=t_i],\\
ACC_{\text{param}}&=\frac{1}{K}\sum_{i=1}^{N_c}\sum_{j=1}^{|p_i|}
\mathbb{I}[|\hat{p}_{i,j}-p_{i,j}|\le\eta]\mathbb{I}[\hat{t}_i=t_i],
\end{aligned}
\end{equation}
where $N_c$ is the sequence length, $t_i$ and $\hat{t}_i$ are the ground-truth and predicted command types, $p_{i,j}$ and $\hat{p}_{i,j}$ are the corresponding parameters, and $K$ denotes the total number of valid parameters in correctly predicted commands. Following DeepCAD, all continuous parameters are quantized into 256 levels, with a tolerance threshold of $\eta=3$.

At the geometry level, predicted CAD sequences are compiled into B-Rep solids and evaluated using Chamfer Distance (CD). Specifically, 2,000 points are uniformly sampled from the surfaces of the reference and reconstructed shapes, denoted as $S_1$ and $S_2$, respectively:
\begin{equation}
CD(S_1,S_2)=
\frac{1}{|S_1|}\sum_{x\in S_1}\min_{y\in S_2}\|x-y\|_2^2+
\frac{1}{|S_2|}\sum_{y\in S_2}\min_{x\in S_1}\|y-x\|_2^2 .
\end{equation}

In addition, we report the Invalid Rate (IR), which measures the proportion of predicted sequences that cannot be compiled into valid B-Rep solids, with $IR=N_{\text{invalid}}/{N_{\text{total}}}$,
where $N_{\text{invalid}}$ and $N_{\text{total}}$ denote the numbers of invalid and total test samples, respectively.

\subsection{Comparison with Other Methods}
\label{subsec:comparision_with_sota}
To demonstrate the superiority of VGGT-CAD, we comprehensively compared our method with two types of state-of-the-art baselines: parametric CAD generation methods and single image-to-3D mesh reconstruction models.

\begin{figure*}[h]
\centering
\includegraphics[width=0.95\textwidth]{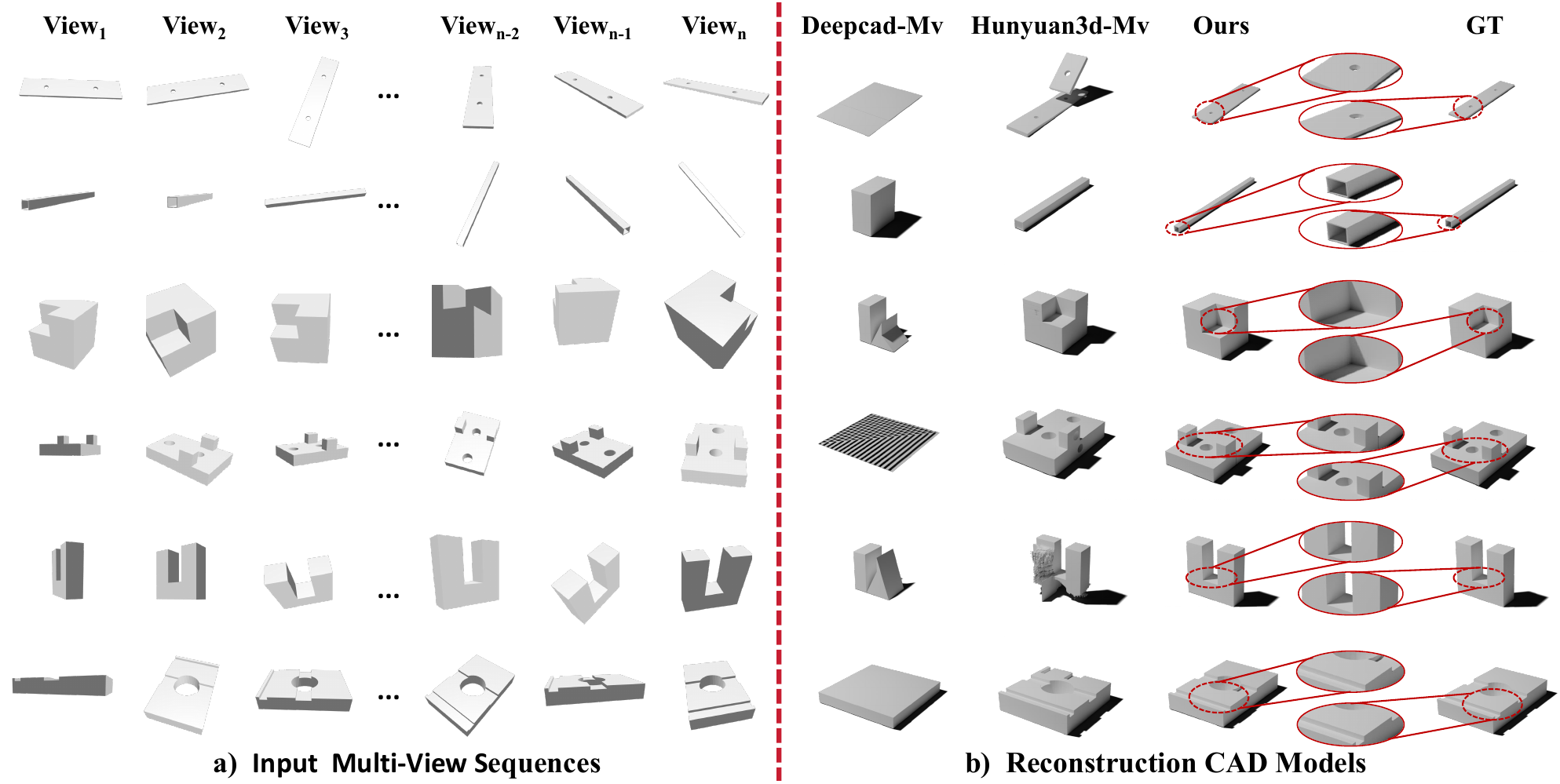}
\vspace{-2.5mm}
\caption{\textbf{Qualitative comparison with existing 3D reconstruction methods.} (a) Input multi-view sequence. (b) Reconstruction results of different methods on objects with diverse geometric structures, including through holes, grooves, and nested components. Red boxes highlight representative regions where Ours better recovers complex back-side topology and fine geometric details, showing closer agreement with the ground truth.}
\label{Fig:Fig_compare}
\vspace{-5mm}
\end{figure*}

\subsubsection{Baselines}
\label{subsubsec:baselines}
In the field of CAD generation, since there is currently no framework specifically designed for ``Multiviews-to-CAD" reconstruction, we modified existing state-of-the-art models to accommodate our input settings. We chose DeepCAD~\cite{DBLP:conf/iccv/WuXZ21} and Hunyuan3d~\cite{DBLP:journals/corr/abs-2501-12202} as our comparison objects. We replaced their original point cloud encoders with our geometric multi-view visual encoder architecture, thereby constructing single-image input variants denoted as DeepCAD* and multi-view video input variants denoted as DeepCAD-Mv and Hunyuan3d-Mv, respectively. Furthermore, in the domain of ``image-to-mesh" reconstruction, we selected TripoSR~\cite{DBLP:journals/corr/abs-2403-02151} and One-2-3-45~\cite{DBLP:conf/nips/LiuXJCTXS23}, which represent the state-of-the-art in fast feed-forward 3D object reconstruction. It should be noted that mesh-based methods cannot output parametric CAD sequences, so they are only evaluated on the Mean CD as geometry-level metric.

\begin{table}[h]
\vspace{-5mm}
\centering
\caption{Comparison of different methods on CAD reconstruction. Higher is better for ACC metrics, while lower is better for CD and IR.}
\resizebox{\linewidth}{!}{%
\begin{tabular}{lcccc}
\hline
\textbf{Method} & \textbf{ACC cmd$\uparrow$} & \textbf{ACC param$\uparrow$} & \textbf{Mean CD$\downarrow$} & \textbf{IR$\downarrow$} \\
\hline
\multicolumn{5}{l}{\textbf{\textit{Image-based Methods}}} \\
DeepCAD* & 0.6511 & 0.6116 & 0.3513 & 0.3427 \\
Hunyuan3d* & -- & -- & 0.4157 & -- \\
TripoSR & -- & -- & 0.7207 & -- \\
One-2-3-45 & -- & -- & 0.1736 & -- \\
Ours* & 0.7854 & 0.6714 & 0.1482 & 0.3535 \\
% Ours* & 0.7634 & 0.6631 & 0.1429 & 0.3694 \\
\hline
\multicolumn{5}{l}{\textbf{\textit{Video-based Methods}}} \\
Hunyuan3d-Mv & -- & -- & 0.4126 & -- \\
DeepCAD-Mv & 0.7509 & 0.6352 & 0.2123 & 0.3826 \\
Ours & \textbf{0.8213} & \textbf{0.7104} & \textbf{0.0670} & \textbf{0.3020} \\
\hline
\end{tabular}
}
\label{tab:cad_results}
\vspace{-2mm}
\end{table}

\subsubsection{Quantitative Results}
\label{subsubsec:quantitative_results}
The quantitative comparison results are shown in Tab.~\ref{tab:cad_results}. VGGT-CAD consistently achieves the best overall performance among the compared methods. Compared with the single-view baselines DeepCAD* and Hunyuan3d*, our method achieves substantial improvements in reconstruction quality, with clear gains in both $ACC_{\text{cmd}}$ and $ACC_{\text{param}}$ over DeepCAD* and a lower Mean CD than Hunyuan3d*. More importantly, multi-view training enables VGGT-CAD to learn implicit structural relationships across viewpoints, allowing the learned geometric priors to compensate for missing or ambiguous structures caused by occlusion and limited visibility. Compared with the multi-view baselines DeepCAD-Mv and Hunyuan3d-Mv, VGGT-CAD further achieves the lowest Mean CD of 0.0670, demonstrating the effectiveness of the proposed camera-conditioned and cross-view modeling design, which is specifically designed to process unconstrained multi-view sequences rather than merely concatenating visual features.

\subsubsection{Qualitative Results}
\label{subsubsec:qualitative_results}
We visualized and compared the 3D reconstruction results of different methods, as shown in Fig.~\ref{Fig:Fig_compare}. Fig. a) displays the input multi-view visualization, covering various viewpoints of the object, while Fig. b) presents a comparison of the reconstruction effects. The test samples range from simple objects of revolution to complex industrial parts with through-holes, grooves, and multi-layered nested structures; meanwhile, the magnified details reveal that our solution retains high fidelity in geometric intricacies.

\subsection{Ablation Studies}
\label{subsec:ab_stud}
We conducted exhaustive ablation studies to verify the necessity of each key design choice in the VGGT-CAD framework.

\subsubsection{Comparison of SFT and LoRA fine-tuning Geometric Multi-View Visual Encoder}
\label{subsubsec:sft_lora}
We first investigated the trade-off between computational efficiency and performance for geometric multi-view visual encoder fine-tuning strategies. As shown in Tab.~\ref{tab:lora_vs_sft}, full supervised fine-tuning (SFT) requires updating a massive 142,699,741 parameters, incurring an extremely high computational cost. By adopting the LoRA fine-tuning strategy, we drastically reduced the number of trainable parameters to only 12,524,822 (an 11.4-fold reduction). Surprisingly, LoRA fine-tuning not only significantly reduced carbon emissions but also achieved performance far exceeding that of full SFT. Specifically, the LoRA variant achieved an $ACC_{\text{cmd}}$ of 0.8213 and reduced the Mean CD to 0.0670, whereas the corresponding metrics for Full SFT were only 0.7202 and 0.2317. We hypothesize that Full SFT led to catastrophic forgetting of the pre-trained visual geometric prior knowledge, while LoRA perfectly preserved the model's powerful generalized visual representation capabilities while effectively adapting to the features of the CAD domain.

\begin{table}[h]
\centering
\caption{Comparison of full fine-tuning (SFT) and LoRA-based tuning. LoRA achieves better performance across all metrics with substantially fewer trainable parameters.}
\resizebox{\linewidth}{!}{%
\begin{tabular}{lcccccc}
\hline
\textbf{Method} & \textbf{ACC cmd$\uparrow$} & \textbf{ACC param$\uparrow$} & \textbf{Mean CD$\downarrow$} & \textbf{Med CD$\downarrow$} & \textbf{IR$\downarrow$} & \textbf{Params (M)} \\
\hline
SFT & 0.7202 & 0.5969 & 0.2317 & 0.0929 & 0.4014 & 142.7 \\
Lora-tuned (Ours) & \textbf{0.8213} & \textbf{0.7104} & \textbf{0.0670} & \textbf{0.0062} & \textbf{0.3020} & \textbf{12.52} \\
\hline
\end{tabular}%
}
\label{tab:lora_vs_sft}
\vspace{-2mm}
\end{table}

\subsubsection{Necessity of Camera Conditioning}
\label{subsubsec:camera_condi}
One of the core innovations of VGGT-CAD is the explicit introduction of camera extrinsic information. To validate this design, we compared the full model with two variants: a baseline model that completely removes camera inputs as ``Baseline", and a variant that only uses camera conditions as encoder inputs without a dedicated Camera Condition Decoder. As shown in Tab.~\ref{tab:ablation_camera}, the baseline model lacking camera inputs struggled to align multi-view features, resulting in a poor Mean CD of 0.1026. Although inputting camera conditions slightly improved geometric accuracy, our full camera-conditioned architecture achieved the best performance across all metrics, reducing the Mean CD by 34.02\% compared to the no-camera baseline. This conclusively proves that explicit camera pose constraints are indispensable for the network to understand 3D spatial transformations between continuous multi-view frames.

\begin{table}[h]
\vspace{-4mm}
\centering
\caption{Ablation study on camera conditioning. Incorporating camera input improves reconstruction quality, while adding camera output further boosts performance.}
\resizebox{\linewidth}{!}{%
\begin{tabular}{lcccc}
\hline
\textbf{Method} & \textbf{ACC cmd$\uparrow$} & \textbf{ACC param$\uparrow$} & \textbf{Mean CD$\downarrow$} & \textbf{Med CD$\downarrow$} \\
\hline
Baseline & 0.8083 & 0.7025 & 0.1026 & 0.0111 \\
+ w/ camera input & 0.8113 & 0.7025 & 0.0742 & 0.0076 \\
+ w/ camera output & 0.8182 & 0.7078 & 0.0677 & 0.0065 \\
+ w/ GAD View Selection & \textbf{0.8213} & \textbf{0.7104} & \textbf{0.0670} & \textbf{0.0062} \\
\hline
\end{tabular}
}
\label{tab:ablation_camera}
\vspace{-4mm}
\end{table}

\subsubsection{Effectiveness of GAD View Selection}
\label{subsubsec:GAD}
To verify the effectiveness of the geometry-aware deterministic view selection module, we replaced the deterministic view selection in the testing phase with random sampling while keeping the network structure, training settings, and viewpoint budget $M$ unchanged. As shown in Tab.~\ref{tab:ablation_camera}, after removing the geometry-aware deterministic view selection, the model experienced a decline across all major metrics. Compared to random sampling, the full model's metrics all improved. These results demonstrate that CAD program recovery depends not only on the modeling capabilities of the decoder but also heavily on the geometric complementarity and visual reliability of the input viewpoint set. The proposed method selects observations with richer structural information under a fixed viewpoint budget by jointly modeling geometric coverage, image quality, and viewpoint redundancy, thereby providing more stable evidence for cross-view aggregation and CAD sequence generation. Since this module does not introduce additional network parameters, the performance boost further proves that deterministic, geometry-aware view selection during testing plays an important role in improving reconstruction accuracy and inference stability.

\begin{table}[h]
\vspace{-4mm}
\centering
\caption{Ablation study of feature aggregation strategies. The Cross-View Context Aggregator outperforms conventional pooling methods in modeling cross-view geometry.}
\resizebox{\linewidth}{!}{%
\begin{tabular}{lccccc}
\hline
\textbf{Method} & \textbf{ACC cmd$\uparrow$} & \textbf{ACC param$\uparrow$} & \textbf{Mean CD$\downarrow$} & \textbf{Med CD$\downarrow$} & \textbf{IR$\downarrow$} \\
\hline
w/ Maxpool & 0.7598 & 0.6596 & 0.1653 & 0.0386 & 0.3631 \\
w/ Meanpool & 0.7626 & 0.6580 & 0.1699 & 0.0353 & 0.3710 \\
w/ Cross-View Context Aggregator & \textbf{0.8213} & \textbf{0.7104} & \textbf{0.0670} & \textbf{0.0062} & \textbf{0.3020} \\
\hline
\end{tabular}
}
\label{tab:ablation_aggregator}
\vspace{-2mm}
\end{table}

\subsubsection{Effectiveness of Cross-View Context Aggregator}
\label{subsubsec:CVCA}
\begin{figure}[h]
\centering
\includegraphics[width=0.5\textwidth]{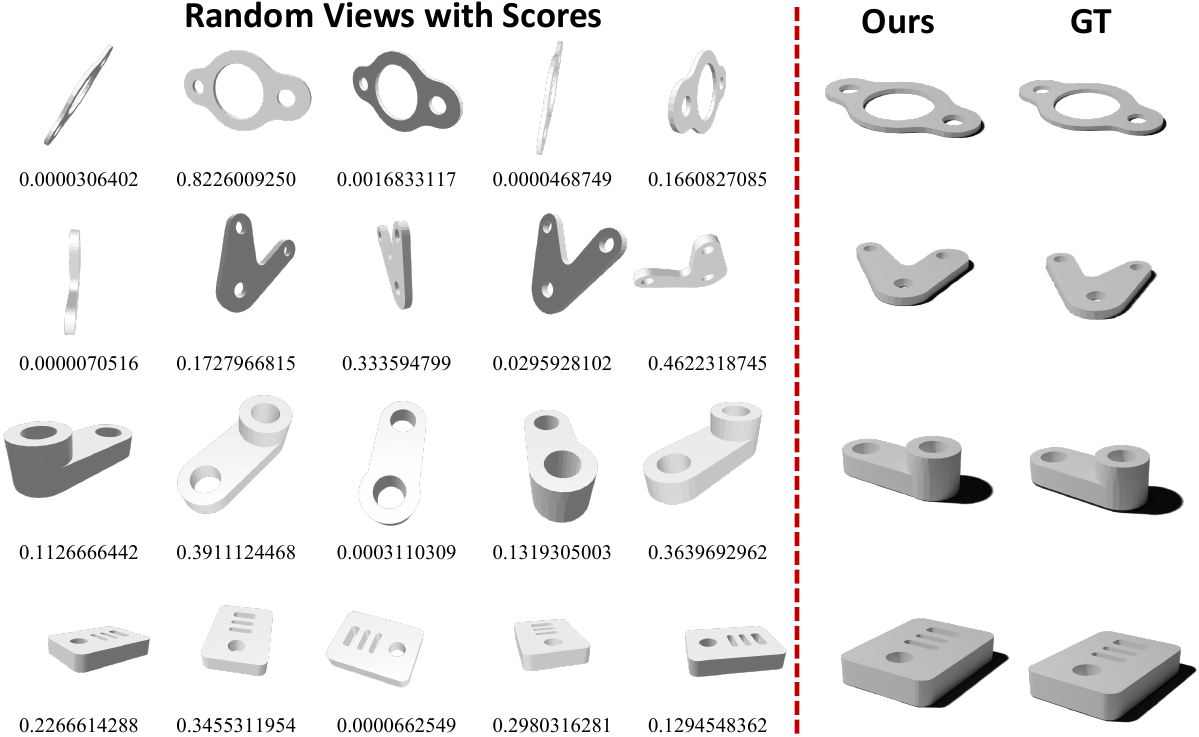}
\vspace{-1mm}
\caption{\textbf{Visualization of cross-view attention weights and reconstruction results.} Left: Candidate viewpoints and their normalized attention weights. Right: Reconstructed CAD models and ground truth. The learned weights favor informative views while suppressing heavily occluded ones, leading to more reliable geometric reconstruction.}
\label{Fig:Fig_select_view}
\vspace{-2mm}
\end{figure}

To evaluate the architectural design of the Cross-View Context Aggregator, we replaced it with standard pooling operations, namely Mean Pooling and Max Pooling, as shown in Tab.~\ref{tab:ablation_aggregator}. Experimental results showed that these traditional pooling strategies failed to effectively integrate cross-view geometric information, resulting in lower command and parameter accuracies. In contrast, our Attention-based aggregator dynamically assigns different fusion weights to each viewpoint based on information density and occlusion status. As shown in the visualizations in Fig.~\ref{Fig:Fig_select_view}, viewpoints capturing complex geometric features are automatically assigned higher attention weights by the network, proving that our aggregator can accurately distill the most valuable geometric cues from lengthy multi-view sequences.

\subsubsection{Effect of Viewpoint Quantity}
\label{subsubsec:views_num}
\begin{figure}[h]
\vspace{-2mm}
\centering
\includegraphics[width=0.5\textwidth]{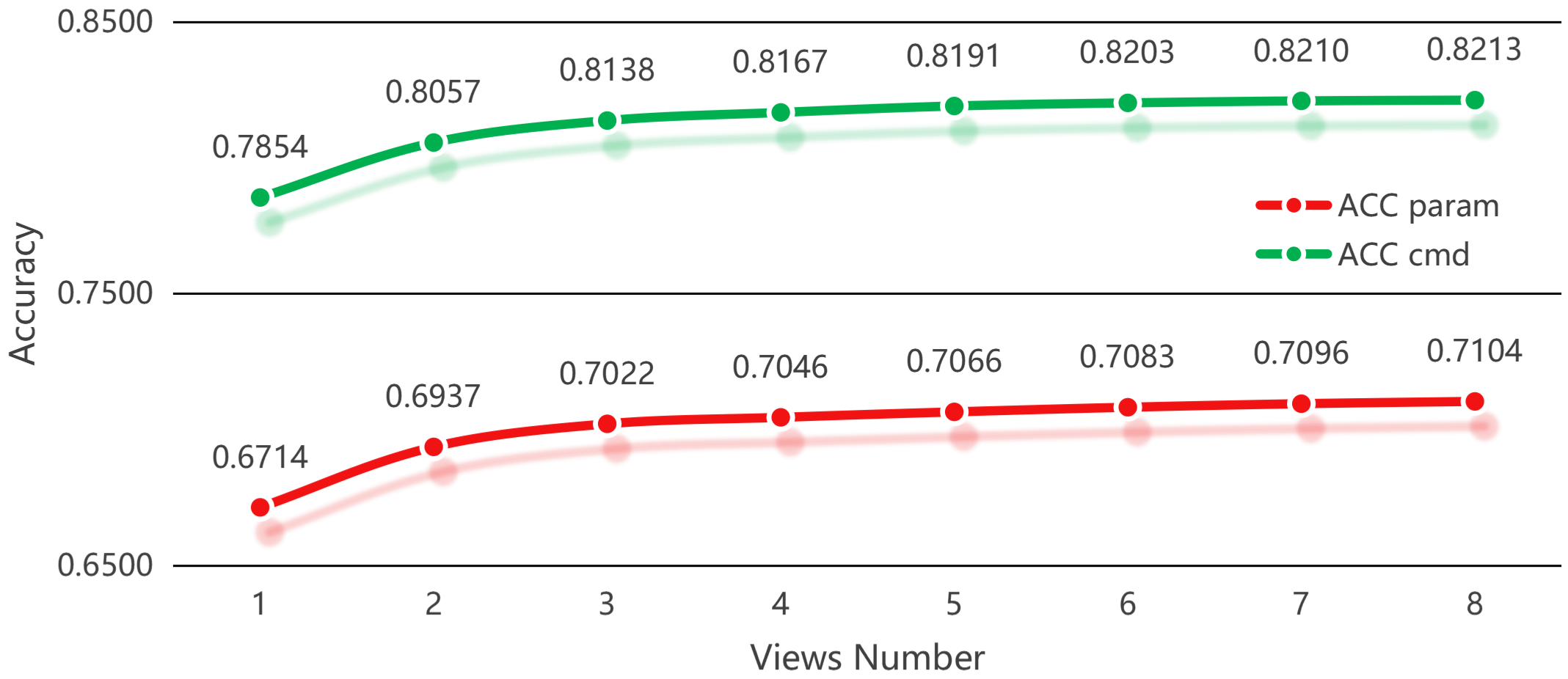}
\vspace{-4mm}
\caption{\textbf{Reconstruction accuracy under different numbers of input views.} The line chart shows $\text{ACC}_{\text{cmd}}$ and $\text{ACC}_{\text{param}}$ consistently improve as more viewpoints are provided, demonstrating the benefit of additional multi-view geometric evidence.}
\label{Fig:views_num}
\vspace{-2mm}
\end{figure}

To investigate the impact of the number of input viewpoints on reconstruction quality, we fixed the model trained on 2 to 8 random views and incrementally increased the number of input viewpoints from 1 to 8 during the testing phase. As shown in Tab.~\ref{tab:multi_view} and Fig.~\ref{Fig:views_num}, when only a single-view input is used, the model's performance is weak due to severe geometric ambiguity. As the number of viewpoints increases, the model can acquire more spatial geometric cues, and the overall trend is improving, but it is not strictly monotonic. Particularly when the number of views increases from 1 to 6, the Mean CD drops significantly; when the number of views reaches 6 to 8, the performance improvement plateaus. This proves the critical role of multi-view inputs in breaking geometric ill-posedness and also demonstrates the model's flexibility in handling variable-length multi-view sequences.

\begin{table}[h]
\vspace{-3mm}
\centering
\caption{Performance under different numbers of input views. More input views consistently improve reconstruction quality and accuracy.}
\resizebox{\linewidth}{!}{%
\begin{tabular}{lcccccccc}
\hline
& \multicolumn{8}{c}{\textbf{Number of Input Views}} \\
\cline{2-9}
\textbf{Metric}
& \textbf{1} & \textbf{2} & \textbf{3} & \textbf{4} 
& \textbf{5} & \textbf{6} & \textbf{7} & \textbf{8} \\
\hline
ACC cmd$\uparrow$   & 0.7854 & 0.8057 & 0.8138 & 0.8167 & 0.8191 & 0.8203 & 0.8210 & \textbf{0.8213} \\
ACC param$\uparrow$ & 0.6714 & 0.6937 & 0.7022 & 0.7046 & 0.7066 & 0.7083 & 0.7096 & \textbf{0.7104} \\
Mean CD$\downarrow$ & 0.1482 & 0.0921 & 0.0761 & 0.0704 & 0.0691 & 0.0694 & 0.0684 & \textbf{0.0670} \\
Med CD$\downarrow$  & 0.0297 & 0.0105 & 0.0079 & 0.0073 & 0.0069 & 0.0069 & 0.0067 & \textbf{0.0062} \\
% IR$\downarrow$      & 0.3535 & 0.3183 & 0.3101 & 0.3035 & 0.3010 & 0.2993 & \textbf{0.2965} & 0.3020 \\
\hline
\end{tabular}
}
\label{tab:multi_view}
\vspace{-2mm}
\end{table}

\subsubsection{Effect of View Selection Terms}

Table~\ref{tab:top_level_view_ablation} evaluates the top-level weights of our deterministic view selection objective. Coverage-only already substantially improves over random sampling, reducing Mean CD from 0.06786 to 0.06505 and improving ACC$_{\text{cmd}}$ from 0.8190 to 0.8198, which indicates that geometric complementarity is the primary driver of effective frame selection. In contrast, quality-only consistently degrades performance, especially on Mean CD and Med CD, suggesting that image quality should be treated as a supportive cue rather than the main selection criterion. The full setting $(\lambda_c,\lambda_q,\lambda_r)=(0.50,0.35,0.15)$ achieves the best ACC$_{\text{cmd}}$ (0.8213), ACC$_{\text{param}}$ (0.7104), IR (0.3020), and Med CD (0.00617), yielding the most balanced overall behavior, even though coverage-only slightly attains the lowest Mean CD.

\begin{table}[h]
\vspace{-3mm}
\centering
\caption{Ablation study of the deterministic view selection terms. The first row uses randomly selected views.}
\label{tab:top_level_view_ablation}
\resizebox{0.5\textwidth}{!}{
\begin{tabular}{ccc|cc|ccc}
\toprule
$\lambda_c$ & $\lambda_q$ & $\lambda_r$
& ACC$_{\text{cmd}} \uparrow$ & ACC$_{\text{param}} \uparrow$
& IR $\downarrow$ & Mean CD $\downarrow$ & Med CD $\downarrow$ \\
\midrule
- & - & -
& 0.8190 & 0.7090
& 0.3027 & 0.06786 & 0.00641 \\
1.00 & 0.00 & 0.00
& 0.8198 & 0.7085
& 0.3024 & \textbf{0.06505} & 0.00619 \\
0.00 & 1.00 & 0.00
& 0.8177 & 0.7060
& 0.3029 & 0.07516 & 0.00726 \\
0.60 & 0.40 & 0.00
& 0.8205 & 0.7095
& 0.3035 & 0.06603 & 0.00621 \\
0.75 & 0.00 & 0.25
& 0.8202 & 0.7098
& 0.3022 & 0.06509 & 0.00618 \\
0.00 & 0.75 & 0.25
& 0.8201 & 0.7082
& 0.3025 & 0.07337 & 0.00684 \\
0.50 & 0.35 & 0.15
& \textbf{0.8213} & \textbf{0.7104}
& \textbf{0.3020} & 0.06695 & \textbf{0.00617} \\
\bottomrule
\end{tabular}
}
\vspace{-2mm}
\end{table}

Table~\ref{tab:primitive_extrusion_ablation} provides a finer-grained evaluation of the same design. Here, Line, Arc, and Circle report exact-match accuracy on the corresponding sketch primitives, i.e., whether the predicted primitive-specific arguments match the ground truth for each curve type; Ext-Plane, Ext-Trans, and Ext-Extent further measure exact-match accuracy on the extrusion parameters for plane orientation, plane translation/size, and extent settings, respectively. Under this evaluation, the full method is consistently strong and achieves the best Arc, Circle, Ext-Plane, and Ext-Trans scores, while coverage-only slightly leads on Ext-Extent. The small margins across settings suggest that the selection objective is robust, with coverage contributing most and redundancy mainly suppressing near-duplicate views without changing the overall trend.

\begin{table}[h]
\vspace{-3mm}
\centering
\caption{Detailed ablation results on primitive recognition and extrusion parameter prediction.}
\label{tab:primitive_extrusion_ablation}
\resizebox{0.5\textwidth}{!}{
\begin{tabular}{ccc|ccc|ccc}
\toprule
$\lambda_c$ & $\lambda_q$ & $\lambda_r$
& Line $\uparrow$ & Arc $\uparrow$ & Circle $\uparrow$
& Ext-Plane $\uparrow$ & Ext-Trans $\uparrow$ & Ext-Extent $\uparrow$ \\
\midrule
- & - & -
& 0.5097 & 0.2415 & 0.5036
& 0.4418 & 0.2781 & 0.3925 \\
1.00 & 0.00 & 0.00
& 0.5105 & 0.2422 & 0.5046
& 0.4461 & 0.2792 & \textbf{0.3978} \\
0.00 & 1.00 & 0.00
& 0.5071 & 0.2371 & 0.4994
& 0.4369 & 0.2769 & 0.3895 \\
0.60 & 0.40 & 0.00
& 0.5108 & 0.2428 & 0.5069
& 0.4465 & 0.2795 & 0.3967 \\
0.75 & 0.00 & 0.25
& \textbf{0.5112} & 0.2425 & 0.5046
& 0.4463 & 0.2790 & 0.3975 \\
0.00 & 0.75 & 0.25
& 0.5082 & 0.2399 & 0.5055
& 0.4424 & 0.2798 & 0.3939 \\
0.50 & 0.35 & 0.15
& 0.5111 & \textbf{0.2434} & \textbf{0.5071}
& \textbf{0.4470} & \textbf{0.2802} & 0.3962 \\
\bottomrule
\end{tabular}
}
\vspace{-4mm}
\end{table}

\section{Conclusion and Limitations}
\label{sec:con_limi} 
We presented \textbf{VGGT-CAD}, a unified framework for end-to-end reconstruction of parametric CAD models from monocular multi-view video sequences with camera information. By coupling camera-conditioned multi-view feature extraction, cross-view context aggregation, and Transformer-based sequence decoding, VGGT-CAD alleviates the geometric ambiguity that limits conventional image-based CAD reconstruction. To support this setting, we also introduced \textbf{VideoCAD}, a large-scale benchmark for multi-view CAD reconstruction. Experiments on both quantitative and qualitative evaluations demonstrate that VGGT-CAD produces more accurate geometry and more faithful CAD command sequences than strong baselines, highlighting the effectiveness of explicit geometric priors for parametric 3D reconstruction.

\noindent\textit{Limitations and Future Work.} Our method remains challenged by thin tubular structures and fully enclosed cavities, whose interior geometry is inherently difficult to infer from visual observations. Future work will explore continuous parameter regression and physics- or manufacturing-aware priors to further improve reconstruction fidelity and practical usability.

\bibliographystyle{IEEEtran}
\bibliography{ref}

\end{document}